\documentclass[runningheads]{llncs}
\usepackage[xindy]{glossaries}

\newcommand{\assert}{\textit{ASSERT}$^{\text{TM}}$}

\newacronym{ltl}{LTL}{Linear-time Temporal Logic}

\newacronym{foltl}{FOLTL}{First-Order Linear-time Temporal Logic}

\newacronym{ctl}{CTL}{Computation Tree Logic}

\newacronym{ptl}{PTL}{Probabilistic Temporal Logic}

\newacronym{pctl}{PCTL}{Probabilistic Computation Tree Logic}

\newacronym{pctl*}{PCTL*}{Probabilistic Computation Tree Logic*}

\newacronym{tctl}{TCTL}{Timed Computation Tree Logic}

\newacronym{stl}{STL}{Signal Temporal Logic}

\newacronym{ltl-x}{LTL-X}{a version of Linear Time Logic without the `next' operator}

\newacronym{mtl}{MTL}{Metric Temporal Logic}

\newacronym{iactlk-x}{IACTLK-X}{a fragment of the temporal-epistemic logic CTLK, without the `next' operator}

\newacronym{mucalc}{$\mu$-calculus}{a strict superset of other temporal logics}

\newacronym{fotl}{FOTL}{First-Order Temporal Logic}

\newacronym{bdi}{BDI}{Belief-Desire-Intention}

\newacronym{prs}{PRS}{Procedural Reasoning System}

\newacronym{are}{ARE}{Autonomous Reasoning Engine}

\newacronym{dmars}{dMARS}{distributed Multi-Agent Reasoning System}

\newacronym{mas}{MAS}{Multi-Agent System}

\newacronym{asm}{ASM}{Abstract State Machines}

\newacronym{tasm}{TASM}{Timed Abstract State Machines}

\newacronym{fsm}{FSM}{Finite-State Machine}

\newacronym{pfsm}{PFSM}{Probabilistic Finite-State Machine}

\newacronym{apfsm}{APFSM}{\textit{Autonomous} Probabilistic Finite-State Machine}

\newacronym{pha}{PHA}{Probabilistic Hybrid Automata}

\newacronym{fsa}{FSA}{Finite-State Automata}

\newacronym{ta}{TA}{Timed Automata}

\newacronym{pta}{PTA}{Probabilistic Timed Automata}

\newacronym{gspn}{GSPN}{Generalised Stochastic Petri Net}

\newacronym{mopn}{MOPN}{Marked Ordinary Petri Net}

\newacronym{cpn}{CPN}{Coloured Petri Net}

\newacronym{ba}{BA}{B\"{u}chi Automata}

\newacronym{dtmc}{DTMC}{Discrete-Time Markov Chain}

\newacronym{ctmc}{CTMC}{Continuous-Time Markov Chain}

\newacronym{mdp}{MDP}{Markov Decision Process}

\newacronym{ioa}{IOA}{Input/Output Automata}

\newacronym{pars}{PARS}{Process Algebra for Robot Schemas}

\newacronym{fsp}{FSP}{Finite State Processes}

\newacronym{piadl}{$\pi$ADL}{$\pi$ calculus combined with ADL}

\newacronym{csp}{CSP}{Communicating Sequential Processes}

\newacronym{hcsp}{HCSP}{Hybrid Communicating Sequential Processes}

\newacronym{shcsp}{SHCSP}{Stochastic Hybrid Communicating Sequential Processes}

\newacronym{ccs}{CCS}{Calculus of Communicating Systems}

\newacronym{wsccs}{WSCCS}{Weighted Synchronous CCS}

\newacronym{csp|b}{CSP$\|$B}{an integrated formalism combining CSP and B}

\newacronym{vhdl}{VHDL}{VHSIC Hardware Description Language}

\newacronym{adl}{ADL}{Architecture Description Language}

\newacronym{slim}{SLIM}{System Level Integration Modelling}

\newacronym{mal}{MAL}{Model Action Logic Language}

\newacronym{acsl}{ACSL}{the ANSI C Specification Language}

\newacronym{fdr}{FDR}{the Failures-Divergences Refinement checker}

\newacronym{jpf}{JPF}{Java PathFinder}

\newacronym{ajpf}{AJPF}{Agent Java PathFinder}

\newacronym{mcmas}{MCMAS}{Model Checker for Multi-Agent Systems}

\newacronym{ltlmop}{LTLMoP}{Linear Temporal Logic MissiOn Planning}

\newacronym{cbmc}{CBMC}{C Bounded Model Checker}

\newacronym{cadp}{CADP}{Construction and Analysis of Distributed Processes}
\newacronym{spear}{SpeAR}{Specification and Analysis for Requirements Tool}

\newacronym{psl}{PSL}{Property Specification Language}

\newacronym{fpga}{FPGA}{Field-Programmable Gate Array}

\newacronym{aadl}{AADL}{Architecture Analysis and Design Language}

\newacronym{lts}{LTS}{Labelled-Transition System}

\newacronym{uml}{UML}{Unified Modelling Language}

\newacronym{sysml}{SySML}{Systems Modelling Language}

\newacronym{robotml}{RobotML}{Robot Modelling Language}

\newacronym{dsl}{DSL}{Domain Specific Language}

\newacronym{sct}{SCT}{Supervisory Control Theory}

\newacronym{sat}{SAT}{Boolean satisfiability problem}

\newacronym{smt}{SMT}{Satisfiability Modulo Theory}

\newacronym{ide}{IDE}{Integrated Development Environment}

\newacronym{mde}{MDE}{Model-Driven Engineering}

\newacronym{ai}{AI}{Artificial Intelligence}

\newacronym{fdir}{FDIR}{Fault Detection, Isolation and Recovery}

\newacronym{ifm}{iFM}{Integrated Formal Methods}

\newacronym{fm}{FM}{Formal Methods}

\newacronym{ros}{ROS}{Robot Operating System}

\newacronym{amcl}{AMCL}{Adaptive Monte Carlo Localisation}

\newacronym{dl}{\text{\upshape\textsf{d{\kern-0.05em}L}}}{differential dynamic logic}

\newacronym{vm}{VM}{Virtual Machine}

\newacronym{bip}{BIP}{Behaviour Interaction Priority}

\newacronym{fol}{FOL}{First-Order Logic}

\newacronym{owl}{OWL}{Web Ontology Language}

\newacronym{ria}{RIA}{Restore Invariant Approach}

\newacronym{ocl}{OCL}{Object Constraint Language}

\newacronym{sil}{SIL}{Safety Integrity Level}

\newacronym{rcl}{RCL}{ROS Contract Language}

\newacronym{rv}{RV}{Runtime Verification}

\newacronym{mc}{MC}{Model Checking}

\newacronym{ssai}{SSAI}{Sub-Symbolic AI}

\newacronym{rml}{RML}{Runtime Monitoring Language}

\newacronym{sats}{SATS}{Small Aircraft Transportation System}

\newacronym{ico}{ICO}{Interactive Cooperative Objects}

\newacronym{gui}{GUI}{Graphical User Interface}

\newacronym{can}{CAN}{Controller Area Network}

\newacronym{pals}{PALS}{Physically Asynchronous, Logically Synchronous}

\newacronym{mtbf}{MTBF}{Mean Time Between Failures}

\newacronym{wcet}{WCET}{Worst-Case Execution Time}

\newacronym{caa}{CAA}{Civilian Aviation Authority}

\newacronym{atc}{ATC}{Air Traffic Control}

\newacronym{dlr}{DLR}{German Aerospace Centre}

\newacronym{assert}{\assert}{Analysis of Semantic Specifications and Efficient generation of Requirements-based Tests}

\newacronym{rae}{RAE}{Requirements Analysis Engine}

\newacronym{esa}{ESA}{European Space Agency}

\usepackage[T1]{fontenc}
\usepackage{hyperref}
\usepackage{graphicx}
\usepackage{color}

\usepackage{svg} 
\usepackage[absolute,overlay]{textpos}

\usepackage{xcolor}
\usepackage{subcaption}

\usepackage{longtable}
\usepackage{multirow}
\usepackage{array}
\usepackage{float}

\usepackage{comment}
\usepackage{orcidlink}

\begin{document}

\title{Revisiting Formal Methods for Autonomous Robots: A Structured Survey}

%
%
\author{Atef Azaiez\inst{1}\orcidlink{0009-0001-5279-686X} \and
David A. Anisi\inst{1}\orcidlink{0000-0003-0870-4259} \and \\
Marie Farrell\inst{2}\orcidlink{0000-0001-7708-3877} \and 
 Matt Luckcuck \inst{3}\orcidlink{0000-0002-6444-9312}}
\authorrunning{A. Azaiez et al.}
%
\institute{Faculty of Science and Tech., Norwegian University of Life Sciences, Ås, Norway \\
\email{\{atef.azaiez,david.anisi\}@nmbu.no}\\
  \and
Department of Computer Science, University of Manchester, Manchester, UK\\
\email{marie.farrell@manchester.ac.uk} \\
\and
School of Computer Science, University of Nottingham, Nottingham, UK
\email {matt.luckcuck@nottingham.ac.uk} 
}

\maketitle              
\begin{abstract}
This paper presents the initial results from our structured literature review on applications of Formal Methods (FM) to Robotic Autonomous Systems (RAS). We describe our structured survey methodology; including database selection and associated search strings, search filters and collaborative review of identified papers. We categorise and enumerate the  FM approaches and formalisms that have been used for specification and verification of RAS. We investigate FM in the context of sub-symbolic AI-enabled RAS and examine the  evolution of how FM is used over time in this field. This work complements a pre-existing survey in this area and we examine how this research area has matured over time. Specifically, our survey demonstrates that some trends have persisted as observed in a previous survey. Additionally, it recognized new trends that were not considered previously including a noticeable increase in adopting Formal Synthesis approaches as well as Probabilistic Verification Techniques.

\keywords{Formal Methods \and Formal Verification \and Formal Synthesis \and Autonomous Robotic Systems \and Survey}
\end{abstract}

\section{Introduction}
Formal Methods have been incorporated in the software production life cycle since the early adoption of computers.
Relying solely on testing has been shown to be not enough to guarantee the absence of bugs in software. This quote from Dijkstra in 1969 emphasises that the computer science community needed to develop alternative methods to testing \textit{"Testing shows the presence, not the absence of bugs"} ~\cite{NATO1970}. 
As technology and aspirations have evolved, the use of Robotic Autonomous Systems (RAS) in  safety- and/or mission-critical applications has increased, including  in the nuclear~\cite{aitken2018autonomous}, aerospace~\cite{webster2011formal}, agriculture~\cite{adam2024case,adam2023case}, transport~\cite{lam2016autonomous} and space domains~\cite{fisher2021overview}. These sorts of critical systems among others involving safety and security requirements clearly need to be robustly verified using Formal Methods for specification and verification. The use of \gls{fm} is admitted, recommended and can be required by some standards ~\cite{LeventiPeetz2025}. The strong verification provided by potentially combinations of distinct \gls{fm} and testing approaches is advantageous as it guarantees mathematical proof of correctness. This aides in the assurance process in critical settings and helps to provide various stakeholders with sufficient confidence that the systems function as expected.
Various\gls{fm} approaches have been developed to fulfil specific needs of verification. 

A 2019 survey provides an overview of these methods and acts as a guidebook for those seeking to apply \gls{fm} in RAS~\cite{10.1145/3342355}. Apart from the obvious benefits to developing reliable software, there is a reciprocal benefit to the \gls{fm} community: the modularity of RAS, as exemplified in the various middleware by which they are supported, fosters creative and interesting opportunities for examining and demonstrating the efficacy of these \gls{fm}~\cite{farrell2018robotics}. These observations have given rise to a novel sub-domain called Formal Methods for Autonomous Systems\footnote{\url{https://fmasworkshop.github.io/}} and many conferences have held special tracks in related topics since. On the other hand, there are some challenges of applying \gls{fm} to RAS, namely the complexity of the this kind of systems as they usually combine discrete software logic with continuous physical dynamics and that can lead to scalability issues. Moreover, the dynamic nature of the environment where RAS operate makes it difficult to capture all interactions and uncertainties. last but not least, there can be a gap between the trustworthiness of formal verification results and the expectations of regulatory acquirements ~\cite{FMAS2021}.   

In this paper, we present the methodology we adopted to conduct our structured literature survey, initial results which examines how the application of \gls{fm} to RAS has evolved over time. We analyse which trends have persisted since the original survey~\cite{10.1145/3342355} and identify emerging trends. We examine the relative use of different formal methods and verification approaches, and discuss the role played by \gls{ssai} (e.g. machine learning). We reflect on potential reasons for these various evolutions, providing insight and set the stage for future development in this field. 

Next, we present some related work, while the rest of the paper is structured as follows. In Sect.~\ref{sec:method} we describe our survey's methodology, including the scope and search terms. We present the results in Sect.~\ref{sec:results} and discuss the implications of the results in Sect.~\ref{sec:discussion}. Finally, Sect.~\ref{sec:conclusion} gives our concluding remarks.
\paragraph{Related Work:}
Our survey builds on a previous survey of \gls{fm} applied to autonomous robotic systems~\cite{10.1145/3342355}, though we extend the time frame from 2007---2018, to 2007---2024; 
We also use a different methodology and work-flow, and used Rayyan~\cite{Rayyan}~\footnote{Rayyan: \url{https://www.rayyan.ai/}} a dedicated tool for conducting structured surveys.
This gives broader coverage of the literature, including both wider search terms and additional search sources, initially returning 20,764 papers.  We examine similar research questions to~\cite{10.1145/3342355} but with an explicit focus on the trends that have emerged over time, and examining the impact of \gls{ssai}.

As the sub-domain of \gls{fm} for autonomous systems has evolved and become more popular over time, it is no surprise that other related research efforts exist. These include a manifesto for applicable \gls{fm} that provides ten principles concerning their use~\cite{gleirscher2023manifesto}. This project however does not specifically focus on RAS, rather it discusses the use and promotion of \gls{fm} in practice more generally.

Leahy et al.~\cite{leahy2024grand} define three categories of grand challenge for verification of autonomous systems: (1) Requirements and Specifications, (2) Models and Abstractions, and (3) Tools, Techniques and Algorithms.
Related work in~\cite{redfield2024verification} provides a research roadmap for verification of autonomous systems which points to several of these open challenges and emerging standards in the area.

\section{Methodology}
\label{sec:method}
\begin{figure*}[t]
 \centering
    \includegraphics[width=\textwidth]{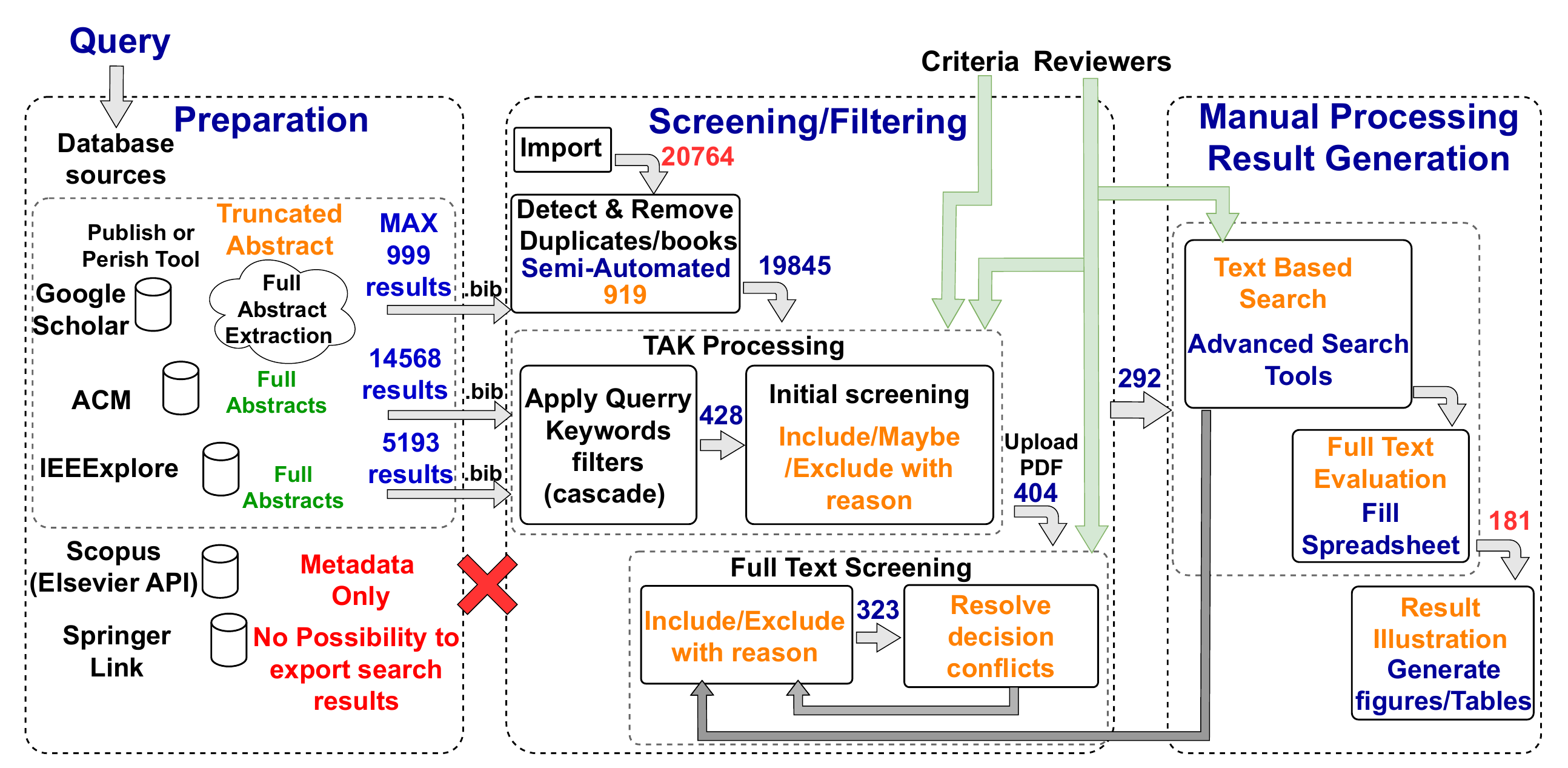}
    \caption{The structured literature review workflow of this paper. The three databases Google Scholar, ACM and IEEExplore were selected as they allowed bulk-export of search results. The Initial query generated 20,764 results. To be able to effectively handle this, they were imported into an online Review Management Platform called Rayyan, to perform initial screening and filtering based on  Title, Abstract, and Keywords (TAK) before continuing with collaborative full text review and result generation.}
    \label{fig:Meth} \label{fig:method}
\end{figure*}

This section outlines the structured methodology, shown in Fig.~\ref{fig:Meth}, that we followed. We begin by defining the relevant Research Questions (\textbf{RQ}) and the scope of the survey. We query three academic search sources using dedicated search strings, designed to gather a broad range of papers. Our initial search produced 20,764 results. To effectively handle this vast number of results, we rely on databases and sources that allow bulk-export of results and import them into an online Review Management Platform, Rayyan, developed for systematic reviews~\cite{Rayyan}. Rayyan supports automated duplicate detection, collaborative initial screening, systematic conflict resolver, and dedicated data summary dashboards. Using Rayyan, we exclude out-of-scope papers using their Title, Abstract, and Keywords (TAK). Then, we import and read the full text of the papers and gather details about the RAS case study and \gls{fm} it uses to answer the \textbf{RQ}s.

\paragraph{Research Questions:}
We define three research questions to direct our search.
     \begin{itemize}
         \item{\textbf{RQ1:}} \textit{What Formal Approaches and Formalisms  are used for specification and verification of RAS?}
         \item{\textbf{RQ2:}}  \textit{Have recent AI advancements within machine-learning affected FM applied to RAS and what FM are used to specify and verify Sub-Symbolic AI (SSAI) enabled RAS?}
         \item{\textbf{RQ3:}} \textit{How has research on FM applied to RAS evolved over time?}
     \end{itemize}

\paragraph{Scope:}
For this survey, we use the following descriptive criteria to define what counts as a RAS, to decide which papers were in-scope for the survey.
\begin{itemize}
     \item RAS are fitted to a physical platform that gives it the capability of navigating in the environment (space/air, ground or water). 
    \item RAS have a certain degree of autonomy, intelligence or adaptability. They may have the ability to adjust their behaviour, or act upon the environment that they operate within, independent of  human interaction.
    \item RAS can consist of multiple agents acting independently or in coordination.
\end{itemize}

\noindent Formal Methods are mathematically rigorous approaches to developing software and systems. They support Specification, Modelling, Design, Synthesis,  and  Verification. As reflected in our search queries, formal verification approaches include theorem  proving, \gls{mc}, and \gls{rv}.

\subsection{Search Strategy}
\begin{figure}[t]
\centering
     \includegraphics[width=\textwidth]{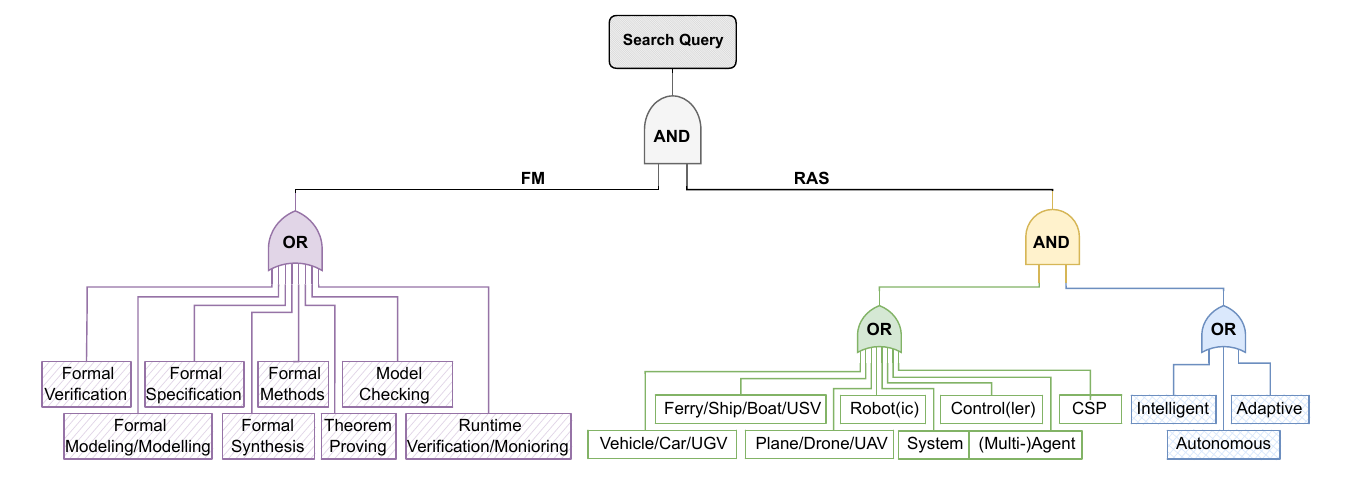} 
    \caption{Query Logical Structure diagram  showing how we use standard logic gate symbols to build our search term from the relevant words and phrases.}
    \label{fig:query}
    \label{fig:qls}
\end{figure}

With the \textbf{RQ}s and scope defined, we formulate the search string(s) that we used to query various data bases. Fig.~\ref{fig:query} shows the Query Logical Structure (QLS)~\cite{9984982} for our search term. It uses standard logic gate operators to illustrate how we combined various words and phrases to build the search term, checking for one word from the \gls{fm} category and one from the RAS category.  

 We use IEEE Xplore, ACM Digital Library and Google Scholar as our search sources as they allow boolean search terms.
 We considered Scopus and Springer Link, but discounted them as they did not support these features.
 Google Scholar does not support exporting search results directly, so we used the Publish or Perish\footnote{Publish or Perish: \url{https://harzing.com/resources/publish-or-perish}} 
 tool to run the Google Scholar search and bulk-export the citations. 
 Each source has a specific query syntax, so we created a distinct search term for each source. In particular, as Google Scholar does not allow more than 256 keywords in the search string, we had to modify and shorten the original search string to remove redundant and less frequently occurring terms. The search terms used for each source are shown below.
 We relied on Rayyan~\cite{Rayyan} to exclude irrelevant papers during the Screening and Filtering process as described in Sect.~\ref{sec:screening}.\\

 \noindent\fbox{\begin{minipage}{\dimexpr\textwidth-2\fboxsep-2\fboxrule\relax}{\small\noindent \textbf{IEEE Xplore:}\textit{ 
((("Full Text \& Metadata": "formal Verification" OR "formal specification" OR "formal modeling" OR "formal modelling" OR "formal synthesis" OR "formal methods" OR "model check*" OR "model-check*" OR "runtime verification" OR "run time verification" OR "run-time verification" OR "runtime monitor*" OR "run time monitor*" OR "run-time monitor*" OR "theorem prov*") AND ("Full Text \& Metadata": "autonomous" OR "intelligent" OR "adaptive")AND("Full Text \& Metadata": "robot*" OR "cyber physical system" OR "CSP" OR "multi*agent" OR "agent" OR "control*" OR system OR vehicle OR car OR UGV OR ferry OR ship OR boat OR USV OR plane OR drone OR UAV)))}
}\end{minipage}}

 \noindent\fbox{\begin{minipage}{\dimexpr\textwidth-2\fboxsep-2\fboxrule\relax}{\small\noindent \textbf{ACM Digital Library:} \textit{AllField:(("formal Verification" OR "formal specification" OR "formal modeling" OR "formal modelling" OR "formal synthesis" OR "formal methods" OR "model check*" OR "model-check*" OR "runtime verification" OR "run time verification" OR " run-time verification " OR "runtime  monitor*" OR "run time monitor*" OR "run-time monitor*" OR  "theorem prov*") AND ( (autonomous OR intelligent OR adaptive) AND (("robot*" OR "cyber physical system" OR "CSP" OR "multi*agent" OR "agent" OR "control*" OR system) OR (vehicle OR car OR UGV OR ferry OR ship OR boat OR USV OR plane OR drone OR UAV))))}
 } \end{minipage}}

  \noindent\fbox{\begin{minipage}{\dimexpr\textwidth-2\fboxsep-2\fboxrule\relax}{\small\noindent \textbf{Google Scholar:} \textit{((formal AND (verification OR specification OR modeling OR modelling OR synthesis OR methods) OR "model*check*" OR ((runtime  OR run*time) AND (verification OR monitor*)) OR  "theorem prov*") AND ( (autonomous OR intelligent OR adaptive) AND (("robot*" OR "CSP" OR "multi*agent" OR agent OR "control*" OR system) OR (vehicle OR car OR plane OR ferry OR drone))))}
}\end{minipage}}\\
~\\ 

Google Scholar only provides truncated abstracts (approx. 30 words) so we manually extracted the full abstracts.
Publish or Perish, however, has a limit of 999 results for bulk-export. Using these sources and search queries, we collected 20,764 results. The majority (14,568) came from ACM Digital Library while the rest (5,193) came from IEEE Xplore.  Next, we describe the process that we followed to screen and filter these papers.
\subsection{Screening and Filtering Workflow}
\label{sec:screening}

We describe our workflow in screening the set of papers produced in the previous step, to remove duplicate results and to filter out papers that are not in scope.
\paragraph{Importing Search Results:}
Rayyan includes advanced features such as automated duplicate detection, collaborative initial screening, systematic conflict resolver, and dedicated data summary dashboards.
\paragraph{Removing Duplicate Papers:}
We identified duplicates using Rayyan's automatic tool by estimating the similarity between papers.
We performed a visual evaluation of potential duplicates with similarity scores below 90 and kept kept the version of the record with the most meta data. 
We finally found 919 duplicate papers, leaving us with 19,841 papers.     
\paragraph{Filtering using Title, Abstract, and Keywords (TAK):} 
To reduce the number of out-of-scope papers that we had to read the full text of, we removed papers where the TAK did not match our scope. We split this process into two stages: (1) an assisted cascade filtering followed by (2) a manual sanity-check and evaluation.
We applied the TAK filtering in a cascading fashion first. We used Rayyan to filter out and exclude papers that did not include any of the \gls{fm} keywords, and exported the remaining set of papers to a new project. In this new project, we applied another filter to exclude papers that did not include any of the Autonomy keywords, and exported this to a second new project.
Finally, we filtered out the papers that did not include any of the Platform keywords. This cascade process produced a set of 428 papers that correspond to our QLS criteria, in Fig.~\ref{fig:method}.
Then, we collectively evaluated the papers produced by the cascade filtering, deciding if they should be included or excluded from the final set of surveyed literature.
Some papers were categorised as ``Maybe'' to be further evaluated based on full text version. After this, 404 papers remained in-scope.

\paragraph{Screening Full Text:} We  uploaded the full text of the papers to Rayyan to resolve the uncertain decisions and provide more context for 
        us as reviewers. After this step, 292 papers were deemed in-scope and processed further manually.

\subsection{Data Collection}   
\label{sec:dataCollection}

\begin{figure}[t]
        \centering 
        \includegraphics[width=0.99\textwidth]{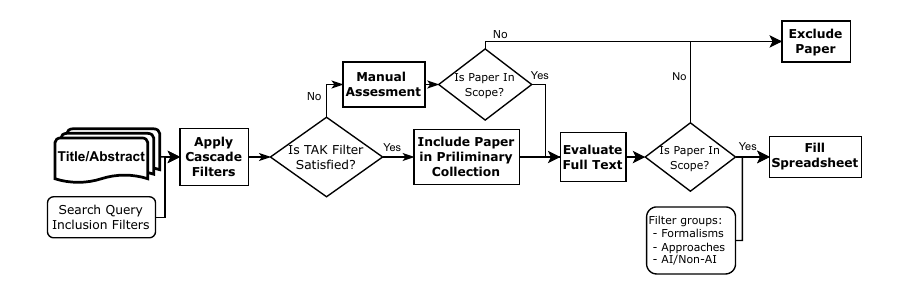}  
               
        \caption{A flowchart of our paper review process. We used the online Literature Review tool, Rayyan, which enabled us to collectively review papers and resolve evaluation conflicts.}
        
        \label{fig:screen_proc}
\end{figure}

We created a spreadsheet to record the information to be collected from the papers.
When reviewing the full text, re-assessment was performed. If any paper was suspected to not meet the inclusion criteria, we flagged it in Rayyan as ``Maybe'' and re-evaluated it collectively. 
The overall review process is shown in Fig.~\ref{fig:screen_proc}. In the end, 181 papers were deemed in scope and used for generating the data and results to answer our \textbf{RQ}s. The complete list of these papers can be found in~\cite{collection}.

\section{Results} 
\label{sec:results}

We present the 
results of the data collection described in Sect.~\ref{sec:dataCollection} and selected surveyed literature in~\cite{collection}. These results underpin the answers to our \textbf{RQ}s presented in Sect.~\ref{sec:method}, showing the spread of different Formal Approaches (Sect.~\ref{sec:approaches}), Formalisms (Sect.~\ref{sec:formalisms}), and \gls{fm} for SSAI(Sect.~\ref{sec:SSAIenabled}) that we found in the surveyed literature. The answers to the \textbf{RQ}s are then collected and made more explicit in Sect.~\ref{sec:RQ_answers}

\begin{figure}[t]
    \centering 
    \includegraphics[width=0.3\linewidth]{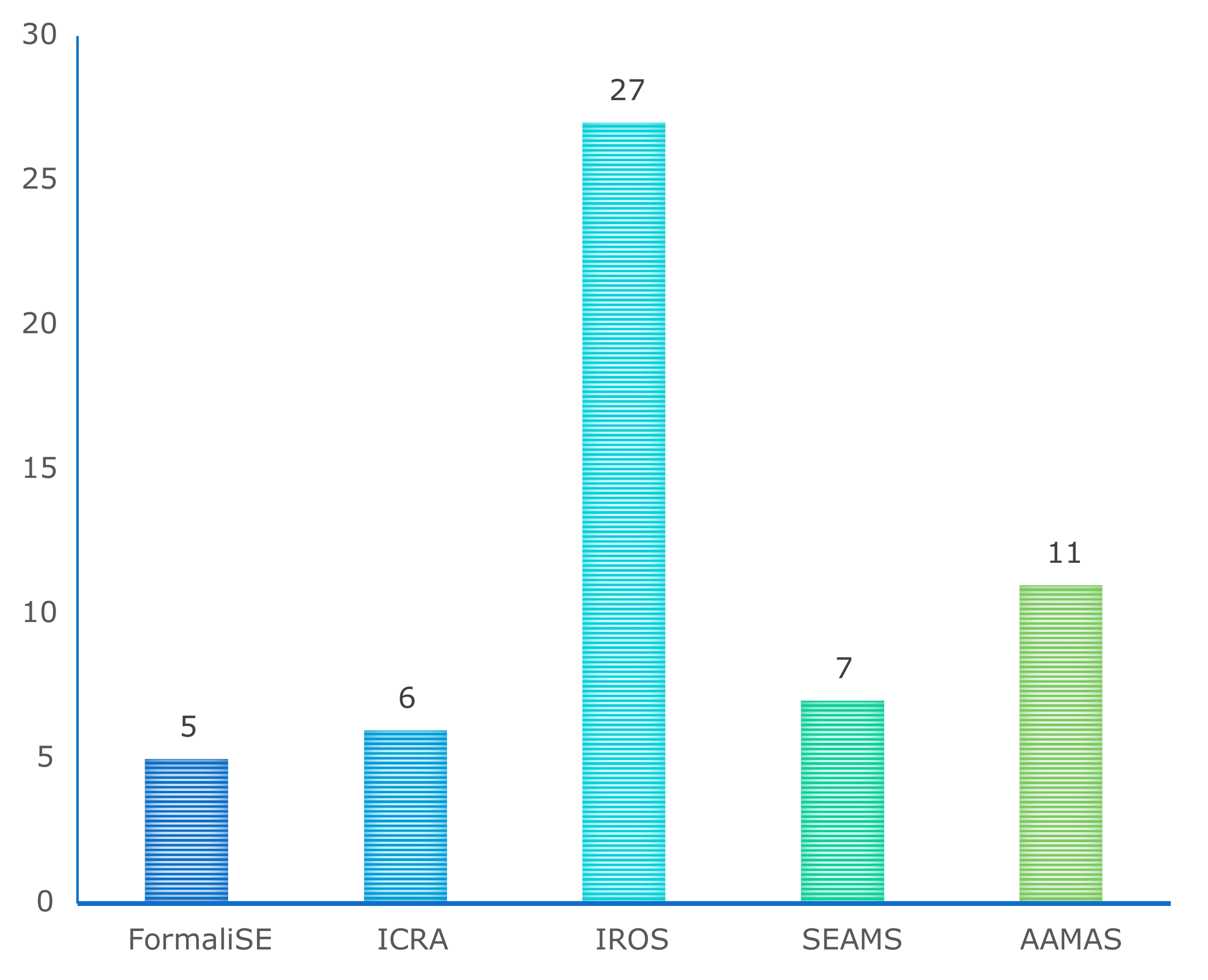} 
        
    \caption{The top 5 venues where the in-scope surveyed literature was published include robotics venues (IROS and ICRA), autonomous systems venues (SEAMS and AAMAS) and a formal methods venue (FormaliSE).}
    \label{fig:venues}
\end{figure}

\subsection{Formal Approaches}
\label{sec:approaches}

\begin{table} [t]
\caption{Formal Approaches}\label{table:approaches}
 \centering
 \scalebox{0.9}{
\begin{tabular}{|l|c|}
  \hline
  \textbf{Approach} & \textbf{Number of papers} \\
  \hline
    Model Checking (MC)       & 130 \\
  \hline
    Runtime Verification (RV) & 24 \\
  \hline
    Theorem Proving      & 18 \\
  \hline
    Formal Plan Synthesis     & 17 \\
  \hline
    Formal Control Synthesis & 15 \\
   \hline
       Formal Specification & 6 \\
  \hline \hline

    Heterogeneous Formal Methods & 31 \\
  \hline
\end{tabular}}
\end{table}
To answer the first part of \textbf{RQ1} - \gls{fm} approaches used for specification and verification for RAS - we counted the number of papers that used \gls{mc}, \gls{rv}, Theorem Proving, Formal Synthesis, and  Formal Specification.
We also counted the papers that used a combination of different \gls{fm} approaches and classified them as Heterogeneous Formal Methods.

Table~\ref{table:approaches} presents the number of papers using each kind of formal approach, showing that \gls{mc} is used 130 times. This concurs with the findings of the previous survey~\cite{10.1145/3342355}.
The approach used the least often  was Formal Specification alone, which we found in only 6 papers in the surveyed literature. Some of these papers focused on formally specifying properties but their mode of verification may have been non-formal, e.g., testing approaches (e.g.,~\cite{rayyan-131658977,rayyan-131658997}). 

\begin{figure}[t]
    \centering
    \begin{subfigure}[b]{0.48\textwidth}
        \includegraphics [width=1\textwidth]{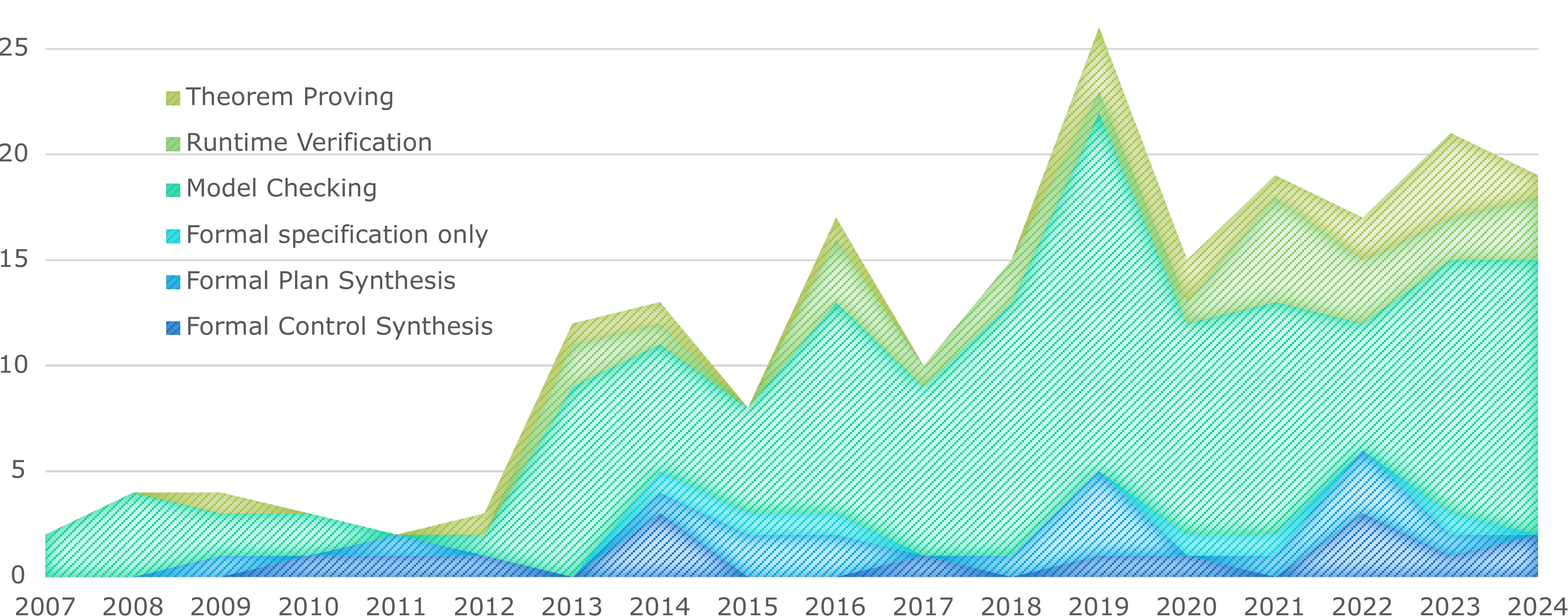}    
        \caption{The number of publications per approach per year.}
        \label{figure:approches_time}
        \end{subfigure}
    \hfill%
    \begin{subfigure}[b]{0.48\textwidth}
        \includegraphics[width=1\textwidth]{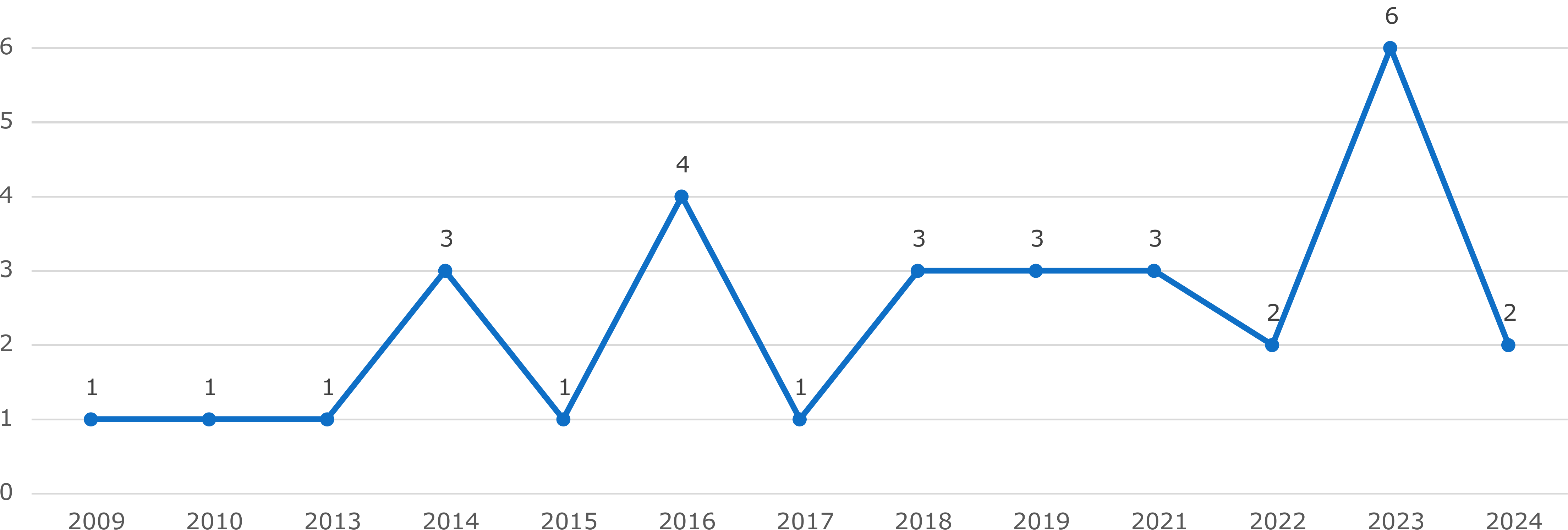}   %
        \caption{The number of Heterogeneous Formal Methods publications identified per year. }
        \label{fig:hetero_time}
    \end{subfigure}
    \caption{Enumerating the number of formal methods approaches per year and, separately, counting the number of heterogeneous approaches per year.}
    \label{figure:Combined_approches_time}
\end{figure}

While 136 out of the 181 papers in the surveyed literature (75\%), used a single formal approach, 31 papers (17\%) used a combination of different, heterogeneous formal approaches.
The use of Heterogenous Formal Methods can be manifested in tight integration between the approaches within a specific component of the RAS (e.g.,~\cite{Rayyan-131659160,Rayyan-131659066}) or separately in different phases of development or components (e.g.,~\cite{Rayyan-131659163}). This result presents an increase in these kind of approaches from~\cite{10.1145/3342355} (up from 13\%) and in at least one paper per year from 2009 to 2024 -- as shown in Fig.~\ref{fig:hetero_time}.
As we shift our focus to FM advancement over time to answer \textbf{RQ3}, another noticeable finding is shown in Fig.~\ref{figure:approches_time} - the number of papers using \gls{rv} has increased substantially from 2013. Also, although theorem proving approaches remain less popular than model checking, Fig.~\ref{figure:approches_time} shows an increase in usage in recent years. This could be due to the inherent limitations of model-checking due to state-space explosion and these theorem proving approaches may have been used to complement other techniques. This trend appears to mirror the increase in Heterogeneous Verification approaches over time shown in Fig.~\ref{fig:hetero_time}.

\subsection{Formalisms}
\label{sec:formalisms}
To answer the second part of \textbf{RQ1} (formalisms used for formal specification and verification of RAS), we quantified the formalisms used (formal language or notations) to specify the system and the property or properties being verified. Most of the papers in the surveyed literature formally specified both the system and properties but some only specified the system (e.g.~\cite{Rayyan-131658965,Rayyan-131658966}) and others only the properties (e.g.~\cite{Rayyan-131658895,Rayyan-131658988}). 
Table~\ref{tab:SystemFormalisms} (resp. Table~\ref{table:PropertyFormalisms}) shows the  number of papers that used the identified formalisms to specify the system (resp. properties) in the surveyed literature. As these tables show, State-Transition formalisms were most often used in the surveyed literature to specify the system, and Temporal Logics were most often used to specify properties. ~\cite{rayyan-131658952} in one example among the 65 papers that used state-transition to model the system and Temporal Logic to model the properties 58 among those papers used Model Checking as an Approach. 
 \begin{table} [t] 
 \begin{minipage}{0.38\textwidth}
     \caption{System Specification Formalisms}\label{tab:SystemFormalisms}
\centering
\scalebox{0.9}{
\begin{tabular}{|l|c|}

  \hline
  \textbf{Formalism (System)} & \textbf{\textnumero~Papers} \\
    \hline
    \hline

    State-Transition        & 117 \\
      \hline
    Differential  Equations & 12 \\
      \hline
    Process Algebra         & 5 \\
  \hline
      Temporal Logic          & 4 \\
  \hline
    Set-Based               & 4 \\
  \hline

    Dynamic Logic           & 3 \\

  \hline

    Formal Ontology                & 3 \\
  \hline
  \hline
    Other                   & 10 \\
  \hline
\end{tabular}}
 \end{minipage}
 \qquad
 \begin{minipage}{0.55\textwidth}
\caption{Property Specification Formalisms}\label{table:PropertyFormalisms}
\centering
\scalebox{0.9}{
\begin{tabular}{|l|l|c|} 
 
  \hline

   \multicolumn{2}{|c|}{\textbf{Formalism (Property)}}  & \textbf{\textnumero~Papers} \\
  \hline
  \hline
    \multirow{7}{*}{Logics }
                            &   Temporal Logic                  & 90 \\ 
                            \cline{2-3}

                            &   Probabilistic Temporal Logic    & 21 \\
                            \cline{2-3}
                            &   Dynamic Logic                   & 4 \\
                            \cline{2-3}
                            &   Other Logics                    & 17     \\
                            \cline{2-3}
                            &\textbf{Total}  & \textbf{132} \\
                            
                            
    \hline
    \multicolumn{2}{|l|}{Process Algebra }                      & 5 \\
    \hline
    \multicolumn{2}{|l|}{Set-Based}                             & 15 \\
    \hline
    \multicolumn{2}{|l|}{Formal Ontology}                              & 3 \\
    \hline
    \hline
    \multicolumn{2}{|l|}{Other}                                 & 17 \\
    \hline

\end{tabular}}
\end{minipage}
\end{table}
\subsection{Formal Methods for Sub-Symbolic AI} 
\label{sec:SSAIenabled}

\begin{figure}[t]
    \centering
    \includegraphics[width=0.48\textwidth]{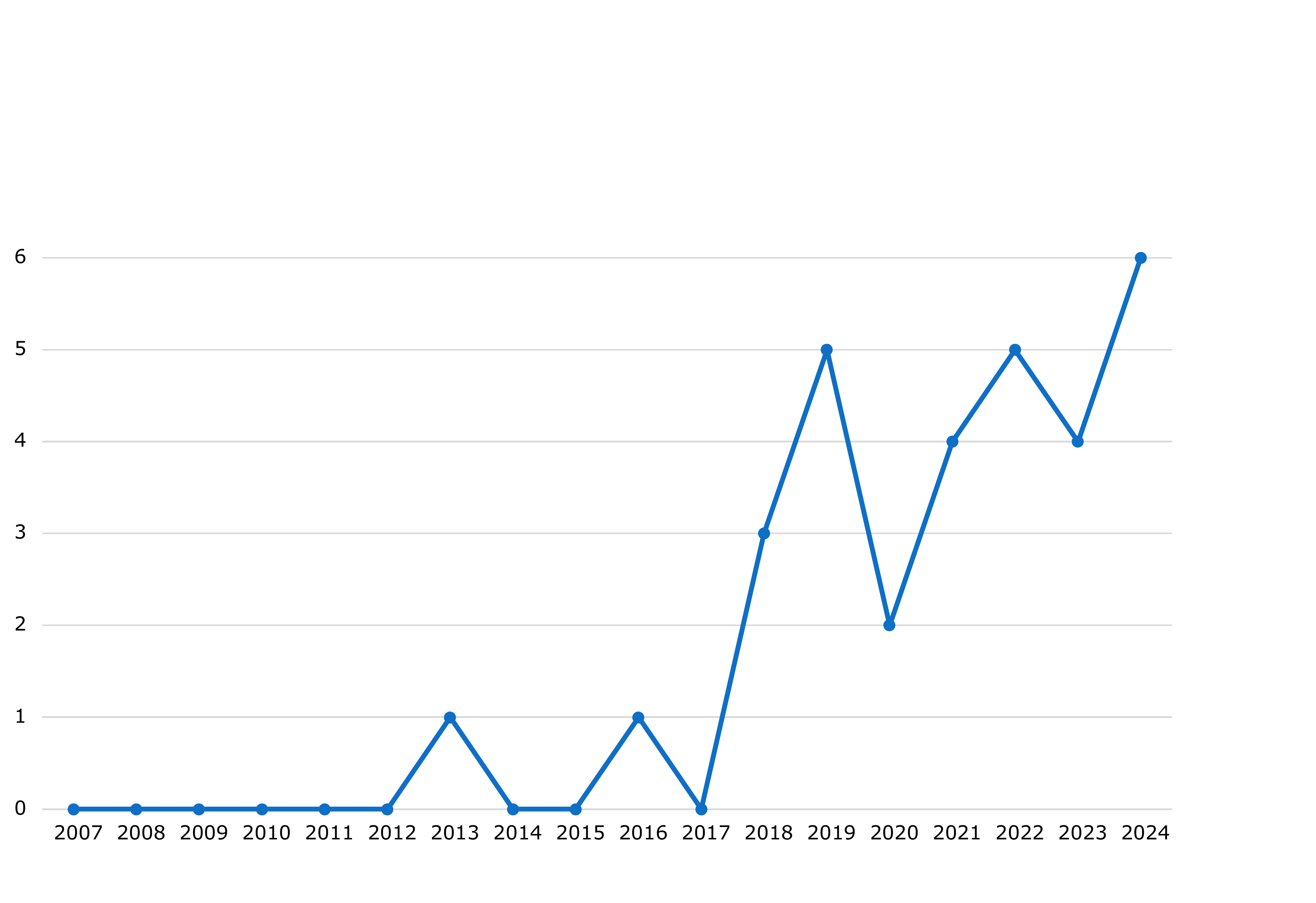}  
    \caption{The number of publications that used FM for \gls{ssai} enabled RAS per year.}
    \label{figure:AI_enabled_trend}
\end{figure}

\begin{figure}[t]
    \centering
    \begin{subfigure}[b]{0.46\textwidth}
    \centering
        \includegraphics[width=\textwidth]{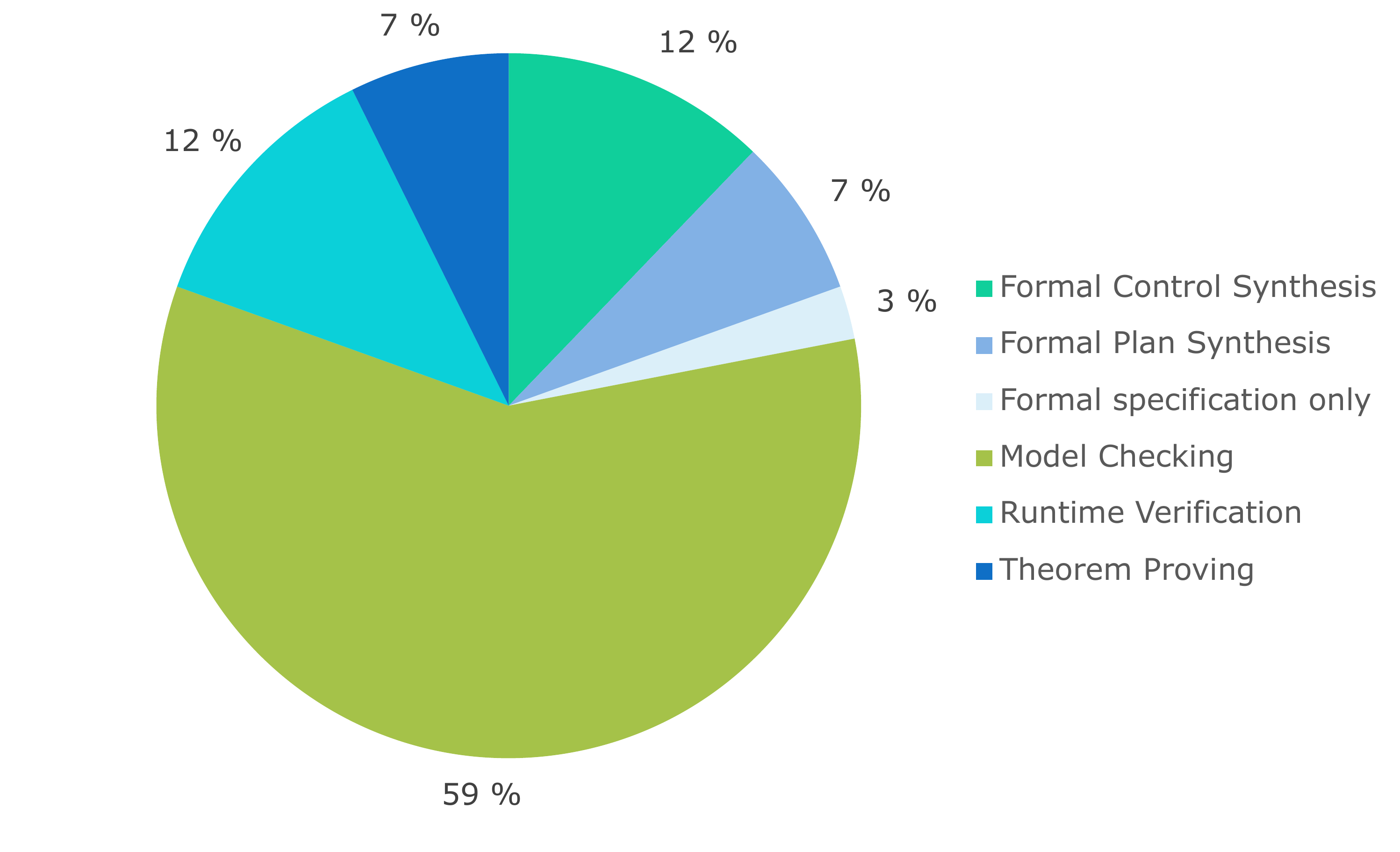} 
        \caption{FM approaches for RAS containing \gls{ssai}}
        \label{fig:pie_ai}
    \end{subfigure}
\noindent    \begin{subfigure}[b]{0.46\textwidth}
    \centering
        \includegraphics[width=\textwidth]{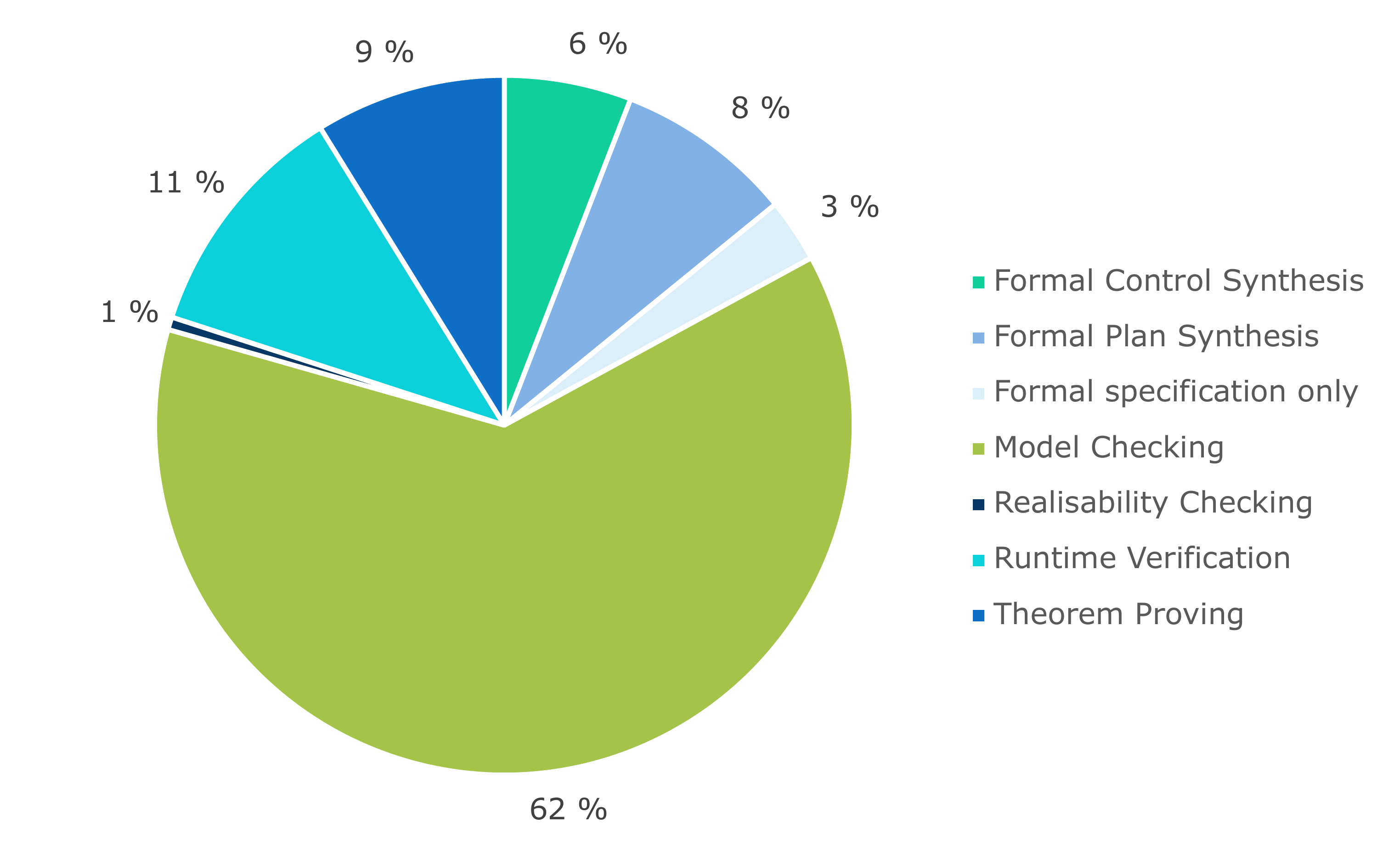}
        \caption{FM approaches for RAS without any \gls{ssai} component}
        \label{fig:pie_non_ai}
    \end{subfigure}
    \caption{Spread of FM approaches for RAS}
    \label{fig:approach_pie}
\end{figure}

To answer \textbf{RQ2}, we highlight the 31 papers from the surveyed literature that use \gls{fm} on at least one \gls{ssai} component. We see an evident trend of increased publications of this kind over the survey period as shown by Fig.~\ref{figure:AI_enabled_trend}.
This could be linked to the recent rise of sub-symbolic AI approaches, most prominently machine-learning, within the survey period.  Fig.~\ref{fig:approach_pie} shows that there is a very similar distribution of approaches applied to RAS using \gls{ssai} versus those papers that don't.


\subsection{Answering the Research Questions} 
\label{sec:RQ_answers}
This section discusses the findings from the results and relates them to our RQs. 

\textbf{RQ1} asks what Formal Approaches and Formalisms are used to specify and verify RAS. We identified a broad range of approaches and formalisms, with Model Checking being the most frequently used approach, and the two most often used formalisms being State-Transition Systems and Temporal Logic. 
The fact that systems were most often specified using State-Transition formalisms is likely a reflection of 88\% of the papers having specified the system using State-Transition formalisms then used \gls{mc} for verification. This pattern is reflected in the property formalisms where Temporal Logics and Probabilistic Temporal Logics are often used alongside State-Transition Systems for MC approaches. Table~\ref{table:PropertyFormalisms} shows that Temporal Logic and its variants are the most used formalism. Here, 76\% of papers that used a Temporal Logic to specify properties, also used a State-Transition formalism to specify the system. Since our survey focuses on RAS, we cannot conclude whether this finding is specific or more common in other domains were \gls{fm} is applied. our interpretation is this could be due to the fact that modelling properties in Temporal Logic is more straight forward to explain to stakeholders, and since RAS typically operate in real-time conditions then modelling the system in State-Transition formalism seem to adequate in several cases.
Note that our survey categorises Probabilistic Temporal Logic and Temporal-Epistemic Logic separately to Temporal Logic, whereas in the previous survey~\cite{10.1145/3342355} these three types of formalism were collected in one single category.
    
\textbf{RQ2} considers the possible effects of recent AI advancements within machine-learning in this domain and asks what FM are used to specify and verify Sub-Symbolic AI (SSAI) enabled RAS.
Although the number of papers demonstrating the use of \gls{fm} in verifying a \gls{ssai} component of RAS is still limited (at most 6 in 2024), we notice an increasing trend from 2018. Most of RAS, if not all, inherently contain one or multiple components that apply an aspect of AI. However, it seems those components start to be more and more critical or involving safety of the assets or humans (e.g. driverless cars, or pilotless aircraft sharing air space with commercial air traffic). Thus, formally verifying these components becomes of greater importance and attracts more research. 
Fig.~\ref{fig:approach_pie} shows, for instance, a similar distribution of \gls{fm} approach distribution except for Formal Control Synthesis that is significantly more used for \gls{ssai} components. This supports the hypothesis that RAS navigation and control algorithms are becoming more and more reliant on \gls{ssai} techniques. This greatly increase the capability of RAS to interact with the environment including humans and therefore calls for the use formal frameworks to guarantee safe and reliable operation. During our survey, we collected more data to study \gls{fm} tools used in the context of \gls{ssai} that will be subject to further analysis. 

\textbf{RQ3} investigates how research on FM applied to RAS has evolved over time.
We focused on the trends and distribution of the various formal approaches throughout the survey period. 
\gls{mc} is the approach most frequently used to specify and verify RAS during the survey period.
The interest in \gls{rv} took some time to appear, but quickly became second. 
This increase could be explained by the time needed for the technology to mature, and the focus was more on the offline verification at early specification and design phases. As RAS become more complex and used in more challenging environments, then \gls{rv} became a valuable framework to ensure that specified properties are met at run-time. Overall, it is reassuring to see the general growth in the use of FM for RAS shown in Fig. \ref{figure:Combined_approches_time}.

\section{Discussion}
\label{sec:discussion}

  In this section, we summarise our threats to validity as follows:

     \paragraph{Threats to Internal Validity:} The automatic duplicate detection and filtering functionality in  Rayyan may have not been as precise as we originally thought. We mitigated against this threat by manually reviewing the papers after the Rayyan classification to minimise any errors/discrepancies.

     The results in Table~\ref{table:approaches} show that most of the papers used either \gls{mc}, \gls{rv}, or Theorem Proving. However this may be a result of having prompted for these three approaches in our searches (Fig.~\ref{fig:qls}). While we also included more general terms, like ``formal verification'' or ``formal specification'', the inclusion of these three specific approaches may have produced search results that are biased to include them over other formal approaches. This means that while we can say that these three approaches were the most often used in the surveyed literature, we cannot say anything about their relative use against other methods that we did not prompt for explicitly. For example, ``Heterogenous Formal Method'' may be less often used in specifying and verifying RAS, or it may simply appear less often in the surveyed literature because of our search terms. 

     \paragraph{Threats to External Validity:} Our search is designed as described in Sect.~\ref{sec:method} to use specific search terms and sources. Both the search terms and sources were broader than in our previous survey~\cite{10.1145/3342355}, aiming to retrieve a more diverse set of results. Hence, comparing the results between these two surveys is not straightforward.
     Further, there is always the possibility that we missed some relevant publications by focusing our search the way that we did. This is reflected in Fig.~\ref{fig:venues} where the most popular venues were in the areas of robotics (IROS/ICRA) and autonomous systems (SEAMS/AAMAS). This is likely a result of searching the IEEE and ACM databases who are the publishers of proceedings at these venues. By excluding Scopus/Elsevier and Springer Link databases, due to their lack of bulk-export capability or ability to provide more than meta-data, papers published in those venues may have only made it into our surveyed literature set, through the Google Scholar search which in turn was channelled through the Publish or Perish import tool and as such constrained to only 999 results. 

     Finally, the fact that each source has a specific query syntax prevented us from using the exact same search term across all three databases. In particular, as Google Scholar does not allow more than 256 keywords in the search string, we had to modify and shorten the original search string to remove redundant and less frequently occurring terms. While we did our best to  minimize the effects of this by carefully selecting and studying the outcome of various changes, it is still noteworthy.

\section{Concluding Remarks}
\label{sec:conclusion}
This paper presents initial results from a structured literature survey on using \gls{fm} to specify or verify RAS. 
 In addition to providing a detailed account of the methodology and work-flow adopted, we considered three Research Questions (RQs), posed and answered in Sect.~\ref{sec:method} and~\ref{sec:RQ_answers}, respectively. 
In particular, we focused on time-evolution of this field, recognising aspects and trends that have not been considered previously, e.g., use of \gls{ssai} methods, Formal Synthesis approaches and Probabilistic Verification Techniques.
 Also, we have identified and currently work on other relevant facets of FM, namely common tools, engineering representations, application domain and multi-agent RAS. A more thorough evaluation of the tools used might also offer more detailed insight into the popularity of the various approaches and formalisms that we found in this survey. 
 These extended results are planned to be part of a journal version of this work.
 
\begin{credits}
\subsubsection{\ackname} This work has received partial funding from the Norwegian Research Council (RCN) RoboFarmer project number 336712, EPSRC grant: EP/Y001532/1 and the Royal Academy of Engineering. 

\subsubsection{\discintname}
The authors have no competing interests to declare that are
relevant to the content of this article.
\end{credits}

\newpage
%
%
%

\clearpage
\bibliographystyle{splncs04}
\bibliography{rayyan-collection-exp, mybibliography }

\end{document}